\documentclass[10pt]{article}
\usepackage{amsmath, amsfonts}
\usepackage{algorithmic}
\usepackage{algorithm}
\usepackage{array}
\usepackage{float}
\usepackage{textcomp}
\usepackage{stfloats}
\usepackage{url}
\usepackage{verbatim}
\usepackage{graphicx}
\usepackage{cite}
\usepackage{pifont}
\usepackage{multirow, multicol}
\usepackage{adjustbox}
\usepackage{hyperref}
\hypersetup{
  colorlinks = true, % Colours links instead of ugly boxes
  urlcolor = blue, % Colour for external hyperlinks
  linkcolor = blue, % Colour of internal links
  citecolor = green % Colour of citations
}

\begin{document}
\begin{center}
{\huge Implement services for business scenarios by combining basic emulators}
\end{center}
\centerline{
Lei Zhao,
Miaomiao Zhang}
\centerline{\textit{China Mobile Research Institute}}
\begin{abstract}

% zh: 本文主要介绍九天智慧网络仿真平台（Jiutian Intelligence Network Simulation Platform）中如何利用各种基础仿真器组合形成组合仿真器，实现对不同业务场景下的仿真服务功能。其中，组合仿真器包括了。而业务场景则包括大话务、多目标天线优化、CSI压缩反馈等不同实际应用情况。

This article mainly introduces how to use various basic emulators to form a combined emulator in the Jiutian Intelligence Network Simulation Platform to realize simulation service functions in different business scenarios. Among them, the combined emulator is included. The business scenarios include different practical applications such as multi-objective antenna optimization, high traffic of business, CSI (channel state information) compression feedback, etc.


\end{abstract}

\section{Introduction}
The Jiutian Intelligence Network Simulation Platform \cite{zhao2023design} decouples, encapsulates and interfaces the key modules of the wireless communication receiving end and transmitting end, and supports module replacement and combination research. Therefore, AI algorithm personnel as users can use the 5G $+$ typical functional network element intelligence and new air interface simulation environment to efficiently and cost-effectively conduct design research and effect verification of new intelligent algorithms in a virtual environment. 

In order to allow users to understand and become familiar with the wireless simulation platform from scratch, especially for algorithm personnel who have no communication system background in the early stage, the smart network simulation platform has designed open tasks (such as multi-objective antenna optimization, high traffic business, CSI compression feedback, etc.) to guide users From familiar use to advanced verification functions.

\section{Combination Emulator}
Individual basic emulators \cite{zhao2023emulators} are responsible for simulating the fundamental capabilities within the network system, but they may not meet the simulation requirements of complex business scenarios. The platform combines multiple basic emulators and provides unified services to meet the task opening and online model training requirements of different business scenarios. Currently, the intelligent network simulation platform provides three types of combination emulators:
\begin{itemize}
    \item \textbf{Real Environment Dynamic User Protocol Stack Simulation}: This combination emulator is composed of user behavior simulation, base station simulation, and terminal simulation. It provides users with metrics such as RSRP, SINR, traffic, and rates at both user and cell granularity in a real environment.
    \item \textbf{Real Environment Dynamic User Coverage Simulation}: To improve the efficiency of base station and terminal simulation, this emulator primarily invokes the physical layer simulation from the above emulators. It combines with user simulation to output coverage metrics such as RSRP and SINR \cite{afroz2015sinr, park2016analysis} at both user and cell levels.
    \item \textbf{Link-Level Channel Simulation}: This combination emulator utilizes the core capability provided by channel simulation. It combines with base station and terminal physical layer protocol stack simulation to provide users with frequency-domain channel response information at the Resource Element (RE) level.
\end{itemize}

% zh: 接下来的文章主要介绍这三个组合仿真器的具体应用场景。
The following article mainly introduces the specific application scenarios of these three combined emulators.

\section{Business Scenarios}
\label{scene}

\subsection{Multi-objective antenna optimization}
% Paragraph 1
% zh: 多目标天线优化任务希望AI模型结合用户行为动态调整天线参数配置，实现网络覆盖及速率性能的协同优化。根据任务开放的可配置参数及参考条件要求，大尺度信道仿真、用户仿真、信道仿真器需要纳入组合；根据任务的仿真结果输出要求，终端/基站仿真器需要纳入组合。图3.5中C1接口为组合仿真器调用接口，通过天线参数配置调用仿真服务，仿真结果通过C2接口输出。组合仿真器内部通过用户仿真器周期性输出用户位置及发起的业务类型，分别输出到大尺度信道仿真与基站仿真；大尺度信道仿真器结合不同的天线配置输入、基站与用户位置、物理环境信息输出衰落模型，通过F1接口与信道仿真器交互；信道仿真结合大小尺度衰落模型计算信道矩阵，通过F4接口输出到基站仿真；基站仿真结合用户业务及信道传播矩阵，估算用户覆盖、干扰及速率性能并组合计算小区级网络性能，最终输出仿真结果。

\begin{figure}[t]
    \centering
    \includegraphics[width=0.6\linewidth]{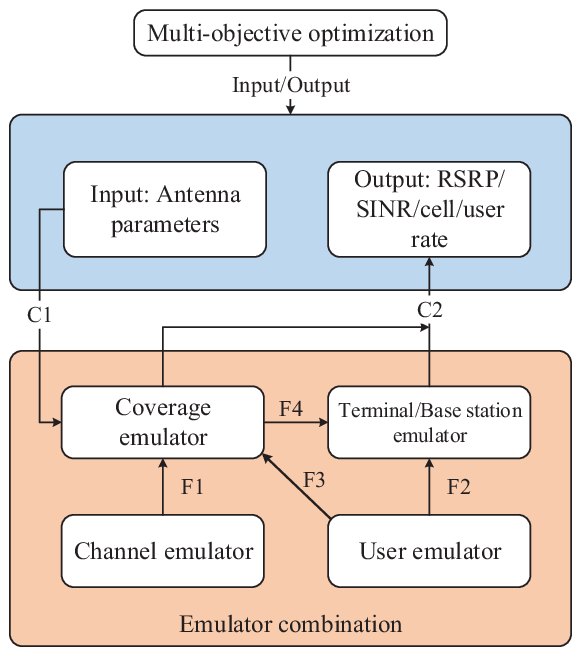}
    \caption{Overview of the combined emulator, in which the definitions of each interface are as follows: C1 is the emulator call, C2 is the emulator output, F1 is the channel model, F2 is the user service, F3 is the user location, F4 is the user RSRP+SINR.}
    \label{fig:multi-1}
\end{figure}

% en:
The multi-objective antenna optimization task aims to have AI models dynamically adjust antenna parameter configurations based on user behavior, achieving collaborative optimization of network coverage and speed performance. According to the configurable parameters and reference conditions specified in the open task, large-scale channel simulation, user simulation, and channel emulator need to be included in the combination. Based on the output requirements of the simulation results for the task, terminal/base station emulators need to be included in the combination.

In Figure \ref{fig:multi-1}, interface C1 serves as the calling interface for the combination emulator, which utilizes antenna parameter configurations to invoke simulation services, and the simulation results are outputted through interface C2. Within the combination emulator, the user emulator periodically outputs user positions and initiated service types, which are respectively fed into the large-scale channel simulation and base station simulation. The large-scale channel emulator, incorporating different antenna configurations, base station and user positions, and physical environment information, outputs a fading model through interface F1 to interact with the channel emulator. The channel emulator, combining large and small-scale fading models, calculates the channel matrix and outputs it to the base station simulation through interface F4. The base station emulator, considering user services and channel propagation matrices, estimates user coverage, interference, and speed performance, combines and computes cell-level network performance, ultimately producing the simulation results.

Reinforcement learning \cite{sutton2018reinforcement, kaelbling1996reinforcement, schulman2017proximal} is applied to multi-objective joint antenna optimization to obtain parameter optimization directions through AI learning, which has been proven feasible in experiments. Below are instructions for the relevant settings.
\begin{itemize}
\item \textbf{Action}:
In multi-objective antenna optimization, the action of reinforcement learning is finally converted into several parameters: horizontal wave width, vertical wave width, azimuth angle, downtilt angle (4 parameters), and whether the beam is effective. However, the implementation of each algorithm may be different. For example, some It may be that the horizontal wave width is adjusted by 1 degree, but it ultimately needs to be converted into a definite value that our base station simulation can identify (historical data is stored according to the 4 parameters that the base station can identify).
\item \textbf{Environment}:
In multi-objective antenna optimization, the environment is the work parameters, maps, user trajectories, user services, base station beam configuration, configuration constraints, base station simulation, user simulation, etc. in the selected optimization area.
\item \textbf{State}:
An environment has several states, and each state is a specific manifestation of the environment. In multi-objective antenna optimization, a state corresponds to a base station distribution in the selected optimization area (generally unchanged), beam configuration (action, variable), user distribution (changes according to time), business model (initially, the service model will be fixed. In the future, the service model of each user will change with time and actions) and network performance (network performance is also a state. It is currently recommended to use performance indicators for 1 second of simulation instead of performance indicators for 5 minutes).
\item \textbf{Reward}:
Based on the action of the previous state and reward output, after acting on the environment, the environment outputs the corresponding network-side performance index, and rewards can be designed based on this performance index. Rewards are calculated by algorithm developers based on performance metrics, using the same source data.
\end{itemize}

\subsection{High traffic business}
High traffic business have always been a more focused scenario in the field of wireless communications. With the emergence of many different scenarios and events such as championships, concerts, and e-sports events, we need simulation capabilities for these specific scenarios. By predicting the traffic pressure around the venue in advance, the communication guarantee and early warning function can be realized.

\begin{figure}[t]
    \centering
    \includegraphics[width=0.95\linewidth]{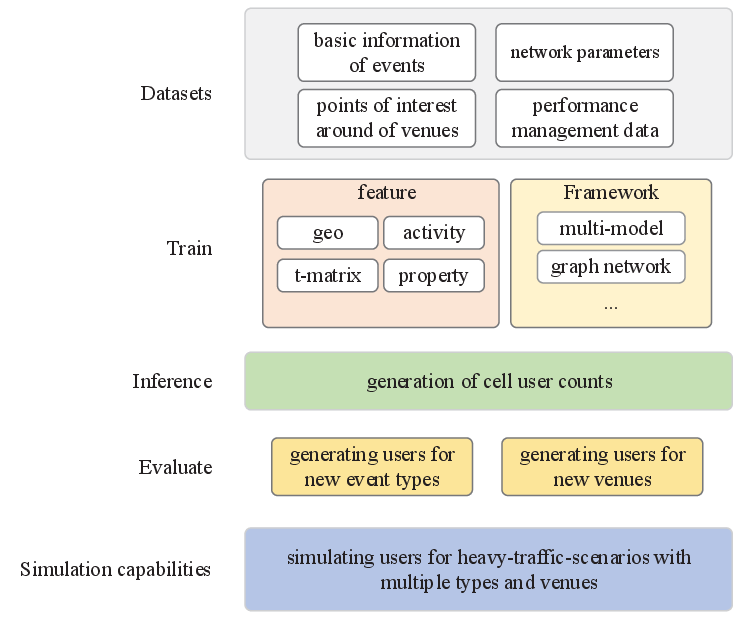}
    \caption{Overview of the high traffic scenario. The graph contains five parts, including dataset, training, inference, evaluation, and simulation capabilities.}
    \label{fig:multi-2}
\end{figure}

As shown in Figure \ref{fig:multi-2}, the main target of the high traffic business is leveraging user location data from high traffic scenarios across different time periods, multiple scenarios, and various event types, to achieve the construction of a generalized and transferable simulation capability for user behavior in these scenarios.

The majority problem of this scenario may be described as utilizing AI capabilities to intelligently generate cell-level user counts for a high traffic scenario from 2 hours before the start to 2 hours after the end, within a radius of 2 kilometers, at a granularity of every 5 minutes. This information is intended to assist in formulating contingency plans and ensuring network security for dual networks.

We proposed two task challenges in this scenario\cite{english1999examining, chen2016xgboost}. $(1)$ Data Scale: The high-traffic events are relatively infrequent, with unpredictable activity times, resulting in a limited amount of data. $(2)$ Portability: Achieving simulation capabilities that can be transferred across venues and different types of events poses a significant challenge. Besides, there are two simplified task to solve its. $(1)$ Starting with congestion prediction, assess the user distribution in the top k cells. $(2)$ Implement migration capabilities in stages, initially focusing on cross-event type migration, followed by cross-scenario migration.

In validation progress, based on data from two concerts, the root mean square error (RMSE \cite{chai2014root}) falls within the range of 11-13, with a mean absolute error (MAE) of 2.8 and a relative error value of 0.42 (relative error value = sum of absolute errors / total number of users). When cross-validated, the results show an RMSE of 28, MAE of 5.957, and a relative error value of 0.54.

\subsection{CSI compression feedback}
\begin{figure}[t]
    \centering
    \includegraphics[width=0.6\linewidth]{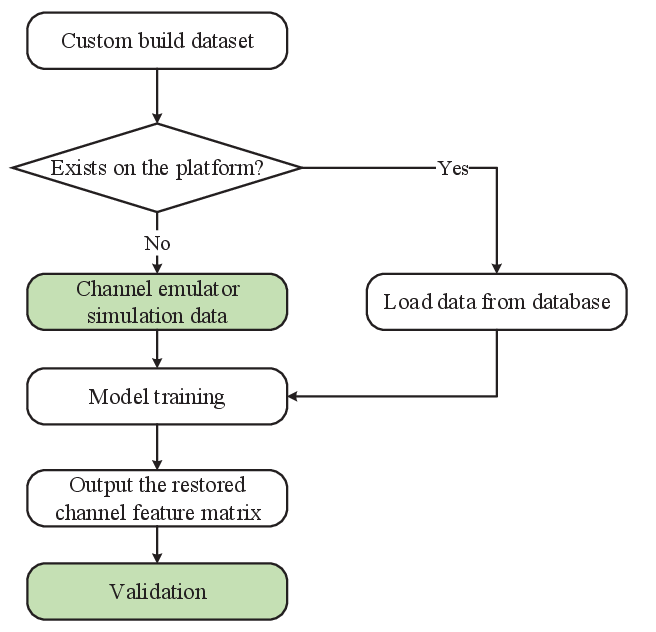}
    \caption{Overview process diagram of CSI.}
    \label{fig:multi-3}
\end{figure}

Relying on the link-level channel simulation of the platform, all functions of the entire process from custom selection of data sets, model training to model effect verification can be realized. The interaction diagram of the CSI compression feedback \cite{guo2022overview} task is shown in Figure \ref{fig:multi-3}.
 
In the task of CSI compressed feedback, the simulation platform capability is called twice. Since this task supports platform users to customize simulation parameters to build datasets, if the platform already has the type of data constructed by this parameter, it only needs to be read directly from the database; If not, you need to call the emulator capability for data simulation. In the restored channel characteristic matrix output from the model training, in addition to considering the evaluation effect of the model indicator NMSE, it is also necessary to call base stations and terminal simulations to perform system verification on the model results. The output business indicators include the total downlink traffic volume of the cell, the average downlink rate of the cell, Average block error rate of the cell.

The task allows users to configure simulation parameters to generate custom datasets. A variety of different scenarios and parameters can be configured to obtain data for model training. The flexible and extensive dataset provides strong data support to solve the problem of model generalization; in addition, the system-level simulation module also conducts a more comprehensive verification of the effect of the model from a perspective closer to the actual network, and provides a model with more reference value. Evaluate effectiveness.

\section{Conclusion}
This article mainly introduces a combination emulator composed of basic emulators in the platform to achieve simulation capabilities for different business scenarios.

Through simulation performance optimization, capability combination and interface encapsulation, the platform provides a good user experience for the external service opening of the simulation network environment and data, and also contributes important infrastructure for the interdisciplinary integration and development of communications and artificial intelligence.

\bibliographystyle{unsrt}
\bibliography{ref.bib}

\end{document}